\documentclass[times,referee,twocolumn,final,authoryear]{elsarticle}

\usepackage{ycviu}
\usepackage{framed,multirow}

\usepackage{amssymb}
\usepackage{latexsym}
\usepackage{makecell}
\usepackage{amsmath}
\newcommand*\rot{\rotatebox{60}}

\usepackage{url}
\usepackage{hyperref}
\usepackage{xcolor}
\definecolor{newcolor}{rgb}{.8,.349,.1}

\usepackage[english]{babel}
\usepackage[utf8]{inputenc}
\usepackage{fancyhdr}

\journal{Computer Vision and Image Understandings}

\begin{document}

\begin{frontmatter}

\title{Human Action Recognition in Drone Videos using a Few Aerial Training Examples}

\author[1]{Waqas \snm{Sultani}\corref{cor1}} 
\cortext[cor1]{Corresponding author: 
  Tel.: +92-3365109108;}
\ead{waqas.sultani@itu.edu.pk}
\author[2]{Mubarak  \snm{Shah}}

\address[1]{Intelligent Machine Lab, Information Technology University, Lahore, Pakistan}
\address[2]{Center for Research in Computer Vision, University of Central Florida, Orlando, USA}

\received{1 May 2013}
\finalform{10 May 2013}
\accepted{13 May 2013}
\availableonline{15 May 2013}
\communicated{S. Sarkar}

\begin{abstract}
Drones are enabling new forms of human actions surveillance due to their low cost and fast mobility. However, using deep neural networks for automatic aerial action recognition is difficult due to the need for a large number of training aerial human action videos. Collecting a large number of human action aerial videos is costly, time-consuming, and difficult. In this paper, we explore two alternative data sources to improve aerial action classification when only a few training aerial examples are available. As a first data source, we resort to video games. We collect plenty of aerial game action videos using two gaming engines. For the second data source,    we leverage conditional Wasserstein Generative Adversarial Networks to generate aerial features from ground videos. Given that both data sources have some limitations, e.g.  game videos are biased towards specific actions categories (fighting, shooting, etc.,), and it is not easy to generate good discriminative GAN-generated features for all types of actions, we need to efficiently integrate two dataset sources with few available real aerial training videos.  To address this challenge of the heterogeneous nature of the data, we propose to use a disjoint multitask learning framework. We feed the network with real and game,  or real and GAN-generated data in an alternating fashion to obtain an improved action classifier. We validate the proposed approach on two aerial action datasets and demonstrate that features from aerial game videos and those generated from GAN  can be extremely useful for an improved action recognition in real aerial videos when only a few real aerial training examples are available.  
\end{abstract}


\end{frontmatter}

\section{Introduction}

Nowadays, drones are ubiquitous and actively being used in several applications such as sports, entertainment, agriculture, forest monitoring, military, and surveillance \cite{Dutta2019AirtoGroundSU_ICRA2019, Cinematography_Action_ICRA2018, MotionCapture_ICRA2018}. In video surveillance, drones can be much more useful than CCTV cameras due to their freedom of mobility and low cost. One critical task in video surveillance is monitoring human actions using drones. 

Automatically recognizing human action in drone videos is a daunting task. It is challenging due to drone camera motion, small actor size, and most importantly the difficulty of collecting large scale training aerial action videos. Computer vision researchers have tried to detect human action in varieties of videos including sports videos \citep{UCF101} , surveillance CCTV videos \cite{cvpr18waqas}, cooking and ego-centric videos \cite{Damen2018EPICKITCHENS}. However, despite being very useful and of practical importance, not much research work is done to automatically recognize human action in drone videos.

Deep learning models are data-hungry and need hundreds of training video examples for robust training. However, collecting training dataset is quite challenging in several vision applications. To address this difficulty of real data collection and its annotations, recently researchers have used games and synthetic images in several computer vision applications such as semantic segmentation \cite{GamePlaying_ECCV}, measuring 6D object pose \cite{ICRA2019_SyntheticPose}, and depth image classification \cite{ICRA2017_SyntheticDepth}. Inspired by the use of video games in  \cite{GamePlaying_ECCV, ICRA2019_SyntheticPose, ICRA2017_SyntheticDepth}, we propose to collect and use game action videos to improve human action recognition in real-world aerial videos. Recently, computer graphics techniques and gaming technology have improved significantly. For example, GTA (Grand Theft Auto) and FIFA (Federation International Football Association) gaming engines use photo-realistic simulators to render real-world environment, texture, objects (human, bicycle, car, etc) and human actions. Games videos for action recognition are intriguing because 1) without much effort, one can collect a large number of videos containing environment and motion that looks close to real-world, 2) It is easy \cite{GamePlaying_ECCV} to get detailed annotations for action detection and segmentation which are otherwise very expensive to obtain, 3) Most of the gaming engines allow the players to simultaneously capture the same action from the different views (aerial, ground, front, etc.,). This means that we can easily collect a large scale multi-view dataset with exact frame-by-frame correspondence. All three advantages make gaming videos quite appealing for aerial action recognition where data collection is difficult and expensive.  To the best of our knowledge, we are the first one to use game videos in aerial action recognition research.  
 \begin{figure}[t]
\begin{center}
   \includegraphics[width=8.9cm,height=3.8cm]{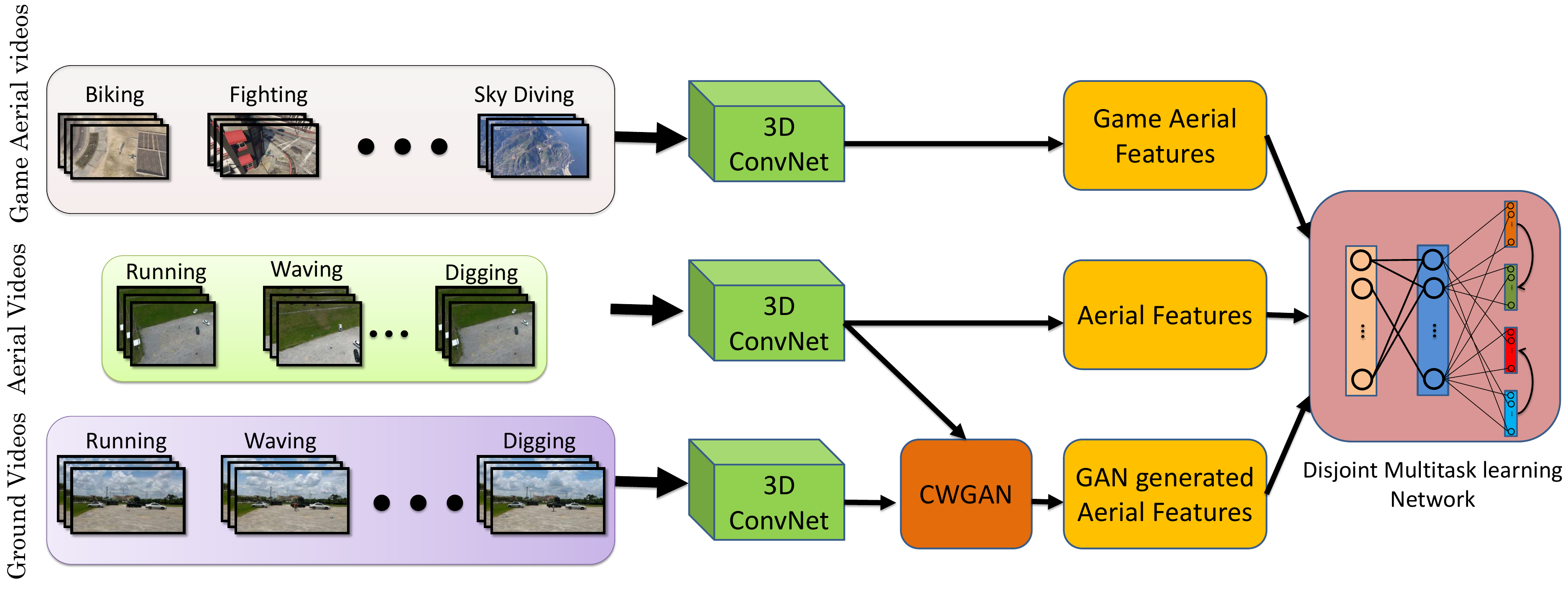}
\end{center}

 \caption{Summary of the proposed training approach. We propose to utilize game videos and GAN generated aerial features to improve aerial action classification when a few real aerial training examples are available. Our approach does not require the same labels for real and game actions. To tackle different action labels in the game and real  dataset, we propose to use disjoint multitask learning framework to efficiently learn robust action classifier.}
\label{fig:Summary}
\vspace{-0.5cm}
\end{figure}
Another direction to address the scarcity of data is to use GAN-generated video examples generated through generative adversarial networks \cite{GAN}. Although the quality of images and videos generated by GAN is not yet good enough to train deep networks\cite{ZeroShot_Xian}, GAN generated discriminative features may be still suitable for action classification. Therefore, we propose to employ conditional Wasserstein GAN \cite{WGAN, CWGAN_CVPR2018} to generate discriminative features.   We believe that the GAN-generated aerial examples, when integrated properly with a few real action examples, can help learn a more generalized and robust aerial action classifier.

In this paper, we propose to utilize game video features and GAN-generated features to improve aerial action classification when a few real aerial training examples are available (see Figure \ref{fig:Summary}). However, one of the key challenges is the disjoint nature of the problem. Video games are designed to address the interest of game playing audience and contain human motions and environments biased towards a few specific human actions. For example, the majority of actions in FIFA games are related to playing a soccer game in a soccer field and the majority of actions in GTA are about fighting. Therefore, it is highly likely that classes of actions in games are different from the types of action classes we are interested to recognize in the real world. Similarly, it is not easy to generate good discriminative GAN-generated features for all types of action. However, our key idea is that despite different classes in games and real videos and the low-quality nature of GAN-generated aerial features, all three data types (games, real and GAN-generated) capture similar local motion patterns, human movements, and human-object interactions, and, if integrated properly, can help learn more accurate aerial action classifiers. To achieve this, we combine games and GAN-generated examples with a few available real training examples using disjoint multitask learning. Specifically, we feed the network with real and game (or GAN-generated) data in an alternating fashion to obtain a more accurate action classifier. Note that in this paper, we call the videos as ground action videos if the person recording the videos is on the ground or at side-angle and the aerial videos are the ones that are taken by UAVs. In summary, this paper makes the following contributions:

$\bullet$ We propose to tackle the new problem of drone-based human action recognition when only a few aerial training examples are available.

$\bullet$ To the best of our knowledge, we are the first one to demonstrate the feasibility of game action videos for improving action recognition in real-world aerial videos. Although game imagery has been used before in different computer vision applications,  it has not been used for aerial action recognition. 

$\bullet$ We show that game and GAN-generated action examples can help to learn a more accurate action classifier through a disjoint multitask learning framework.  

$\bullet$ We present two new action datasets: 1) Aerial-Ground game dataset containing seven human actions where for each action we have 100 aerial-ground video pairs, 2) Real aerial dataset containing actions corresponding to eight actions of UCF101.

 \section{Related Work}

Human action recognition in videos is one of the most challenging and active vision problems \cite{PAMI_Saadali,IDTF,Sultani_2014_CVPR,I3D,C3D,Omar_Oreifej_ICCV11,TwoStream,multifiber,ICRA_Action_Context2019,Cinematography_Action_ICRA2018,MotionCapture_ICRA2018}. Classical approaches used hand-crafted features \cite{PAMI_Saadali,IDTF} to  train generalized human action recognition models that can perform well across different action datasets \cite{Cross-Dataset,Sultani_2014_CVPR}.

With the resurgence of deep learning, several deep learning approaches have been proposed for action recognition. Simonyan et al. \cite{TwoStream} proposed  RGB and optical flow-based networks for action recognition videos. Both RGB and optical flow networks employ 2D convolution. Tran et al.  \cite{C3D} demonstrated the feasibility of 3D convolution for action recognition. In addition to presenting a new large scale action recognition dataset of 400 classes, Carreira et al. \cite{I3D} proposed a two-stream inflated 3D ConvNet (I3D) that is based on 2D convnet inflation and demonstrated state of the art classification accuracy. Recently, an efficient action recognition framework is proposed by Chen et al. \cite{multifiber}. Furthermore, there has been an increased interest to train the generalized action recognition model using multi-task learning.  Kataoka et al. \cite{ICRA_Action_Context2019} put forwarded a multi-task approach for the out-of-context action understanding. Similarly, Kim et al. \cite{DisJMTL} proposed disjoint multi-task learning to obtain improved video action classification and captioning in a joint framework.

Recently, Zhou et al. \cite{MotionCapture_ICRA2018} proposed to analyze human motion using videos that are captured through a drone that orbits around the person. They demonstrated that, as compared to static cameras, videos captured by drones are more suitable for better motion reconstruction. Similarly, Huang et al. \cite{Cinematography_Action_ICRA2018} presented a system that can detect cinematic human actions using 3D skeleton points employing a drone.

Although human action recognition is quite an active area of research in computer vision, there does not exist many research works in the literature that deals with aerial action recognition. Wu et al., \cite{Omar_Oreifej_ICCV11} proposed to use low-rank optimization to separate objects and moving camera trajectories in aerial videos.  UCF-ARG dataset \cite{UCFARG} contains ground, rooftop, and aerial triplets of 10 realistic human actions. This dataset is quite challenging as it contains severe camera motion, non-discriminative backgrounds, and humans in these videos occupy only a few pixels.  
Perera et al. \cite{UAV-Gesture-dataset} proposed to use human pose features to detect gestures in aerial videos. They introduced a dataset that is recorded by a slow and low-altitude (around 10ft) UAV. Although useful, their dataset only contains gestures related to UAV navigation and aircraft handling. Recently,  Barekatain et al. \cite{Okutama-Action} proposed a new video dataset for aerial view concurrent human action detection. It consists of 43 minute-long fully-annotated sequences with 12 action
classes. They used a single-shot detection approach \cite{SSD} to obtain human bounding boxes and then used features within those bounding boxes for action classification. They have neither addressed the problem for the less number of training videos nor they have used the multiple data sources.

Gathering large-scale datasets and its annotation is expensive and requires hundreds of human hours. To address this challenge, there is an increasing interest in employing synthetic data to train deep neural networks. Josifovski et al. \cite{IROS_SyntheticObjectDetection} proposed to use annotated synthetic data to train the instance-based object detector and 3D pose estimator. Mercier et al. \cite{ICRA2019_SyntheticPose} used weakly labeled images and synthetic images to train a deep network for object localization and 6D pose estimation in real-world settings. Carlucci et al. \cite{ICRA2017_SyntheticDepth} proposed to use synthetic data for depth image classification. Recently, Richter et al., \cite{GamePlaying_ECCV} designed a method to automatically gather ground truth data for semantic segmentation and \cite{Hong_2018_CVPR} presented a GAN based approach to use game annotations for semantic segmentation in real images. Finally, Mueller et al., \cite{uav_benchmark_simulator_ECCV} put forwarded photo-realistic simulators to render real-world environment and provide a benchmark for evaluating tracker performance.

In this paper, in contrast to the above-mentioned methods, we demonstrate the feasibility of game action videos for improving action recognition in real-world \textit{aerial} videos. To evade collecting costly drone training videos, we claim to provide a unified framework employing games and GAN-generated data to achieve improved aerial action recognition. No one has used disjoint multitask learning for aerial action recognition or with three different data types. Note that although we use real ground videos for GAN-generated features, our approach does not require exact ground-aerial pairs. Furthermore, our game dataset collection highlights the built-in multi-view action capturing feature in games that can be used for multi-view action recognition. Multiple views make the dataset more extensive and open avenues for other researchers to solve novel challenging problems. We believe that our dataset will push the research in joint game-real aerial action recognition.

\section{Proposed Approach}
 
In this section, we provide the details of our game actions video collection, the method to generated GAN-generated features, and finally disjoint multitask approach where we train the aerial action classifier using game aerial, GAN-generated aerial, and a few real aerial videos in a unified framework. 

\subsection{Games Action Dataset}\label{Games Dataset}
 
\begin{figure}[h!]
\begin{center}
   \includegraphics[width=8.8cm,height=10cm]{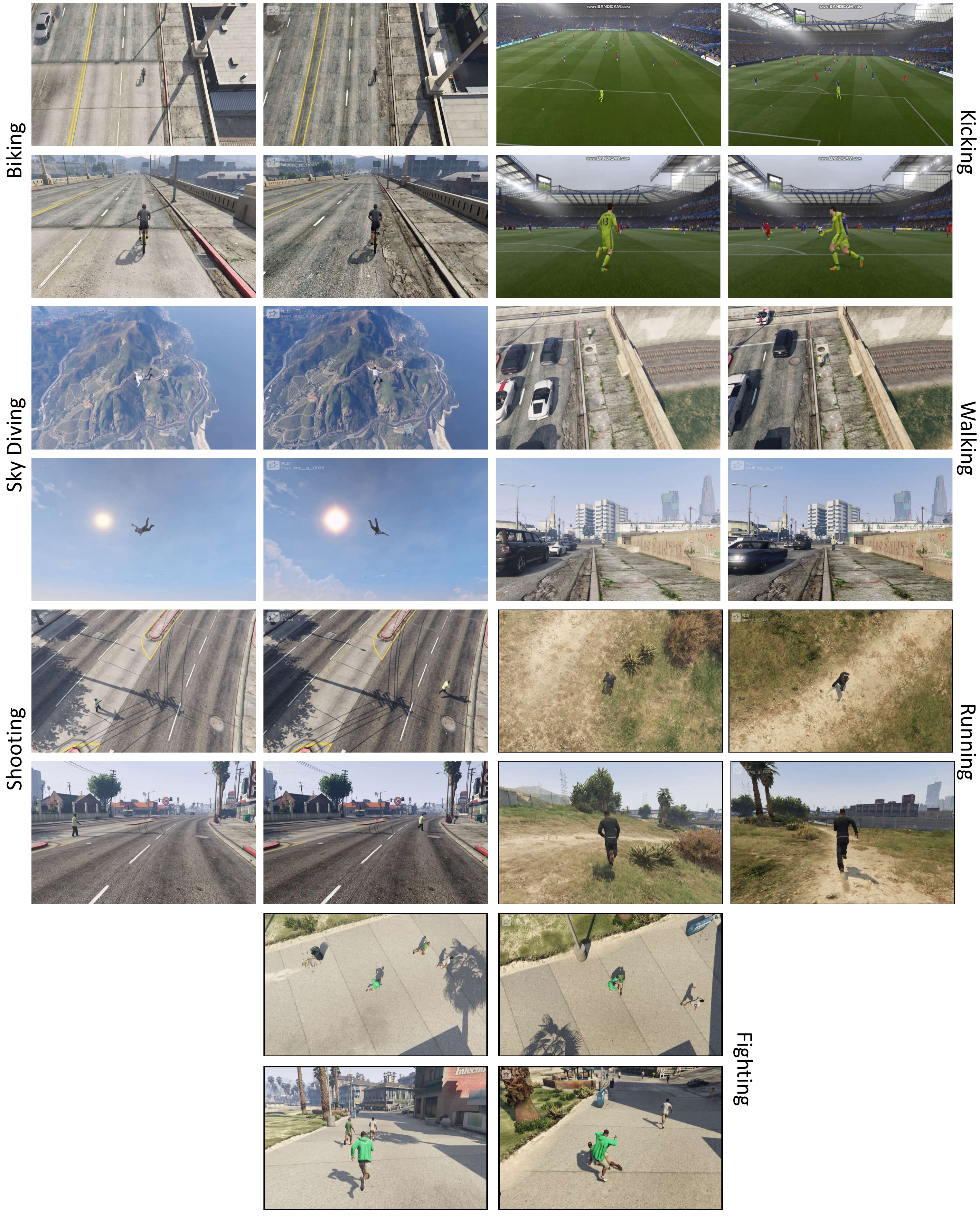}
\end{center}
 \caption{Two frames of each action for both aerial and ground views from our game action dataset. The first, third, fifth and seventh row represent aerial videos and second, forth, sixth, and eight row shows the ground videos. For each action, we show the two frames per video.}
\label{GameDataset2_compressed}
\end{figure}
We employ GTA-5 (Grand Theft Auto) and  FIFA (Federation International Football Association) for collecting the game action dataset \footnote{The customized engine such as \cite{Vivid} can also be used to collect more game videos}. We ask the players to play the games and record the same action from multiple views. Note that GTA and FIFA allow users to record the actions from mutiple angles with real-looking scenes and different realistic camera motions.
 In total, we collect seven human actions including cycling, fighting, soccer kicking, running, walking, shooting, and skydiving. Due to the availability of plenty of soccer kicking in FIFA games, we collect kicking from FIFA and the rest of the actions are collected from GTA-5. Although in our current approach we are only using aerial game video, for more complete dataset purposes, we capture both ground and aerial video pairs i.e., the same action frames captured from both aerial and ground cameras. Figure \ref{GameDataset2_compressed} shows two frames of each action for both aerial and ground views. These videos will be made publicly available.

 For each action, our dataset contains 200 videos (100 ground and 100 aerial) with a total of 1400 videos for seven actions. Note that most of the scenes and interactions in the video games are biased towards actions related to fighting, shooting, walking and running, etc. Therefore, employing game videos to improve action recognition in real-world videos is not trivial. Therefore, in this paper, we propose a  unified approach to combine games and real videos employing disjoint multitask learning.

\subsection{GAN-generated Aerial Examples} \label{WGAN_Section}
We generate GAN-generated aerial video features using Generative Adversarial Networks (GAN) \cite{GAN}.
GAN consists of two networks: Generator and Discriminator. Generator tries to mimic the real data distribution and fools the discriminator by producing realistic looking videos or features while the discriminator job is to robustly classify real and generated video or features. Both generator and discriminator can be simple multi-layer perceptrons. As compared to vanilla-GAN, in conditional GAN \cite{CGAN}, both generator and discriminator are conditioned on auxiliary information. Auxiliary information can be video labels or some other video features. Our goal is to generate GAN-generated aerial visual features given the real ground features (auxiliary information). Therefore, in our case, the objective function of conditional GAN is given by:
\begin{align}
\begin{split}
\mathcal{L}_{cgan} &=  \mathbb{E} [log D(f_{r_a}|f_{r_g})] \\
&+ \mathbb{E} [log(1- D(G(z,f_{r_g})|f_{r_g}))],
\end{split}
\end{align}
where $D$ represents discriminator and $G$ represents generator, in $D(f_{r_a}|f_{r_g})$, $f_{r_a}$ and $f_{r_g}$ are real aerial and ground features respectively. These features are randomly sampled from given real aerial and ground features distributions. Note that we do not assume any correspondence between $f_{r_a}$ and $f_{r_g}$. Given the noise vector $z$ and $f_{r_g}$, generator tries to fool discriminator by producing GAN-generated aerial features. 

To optimize the above objective function, usually KL or JS divergence is employed to reduce the difference between real and generated data distributions. However, one of the key limitations with KL or JS divergence is that the gradient of divergence decreases with the increase of distance, and the generator learns nothing through gradient descent. To address this limitation, recently Wasserstein GAN is introduced in \cite{WGAN, CWGAN_CVPR2018}, which uses Wasserstein distance. WGAN learns better because it has a smoother gradient everywhere. Finally, to make Wasserstein distance tractable, the 1-Lipschitz constraint is used through gradient penalty loss \cite{WGANGP}. The objective function of generating GAN-generated aerial features using conditional Wasserstein GAN (WCGAN-GP) is given by:
\begin{align}
\begin{split}
\mathcal{L}_{cwgan} &= \mathbb{E} [D(G(z,f_{r_g})|f_{r_g})]-\mathbb{E} [log D(f_{r_a}|f_{r_g})] \\
&+ \mathbb{E} [(\left \|\nabla_{m} D(m,(G(z,f_{r_g}))\right \|_{2}-1)^2],
\end{split}
\end{align}
where $m=tG(z,f_{r_g})+ (1-t)f_{r_g}$ and $t$ is uniformly sampled between 0 and 1.
\begin{figure}
\begin{center}
   \includegraphics[width=8.8cm,height=4cm]{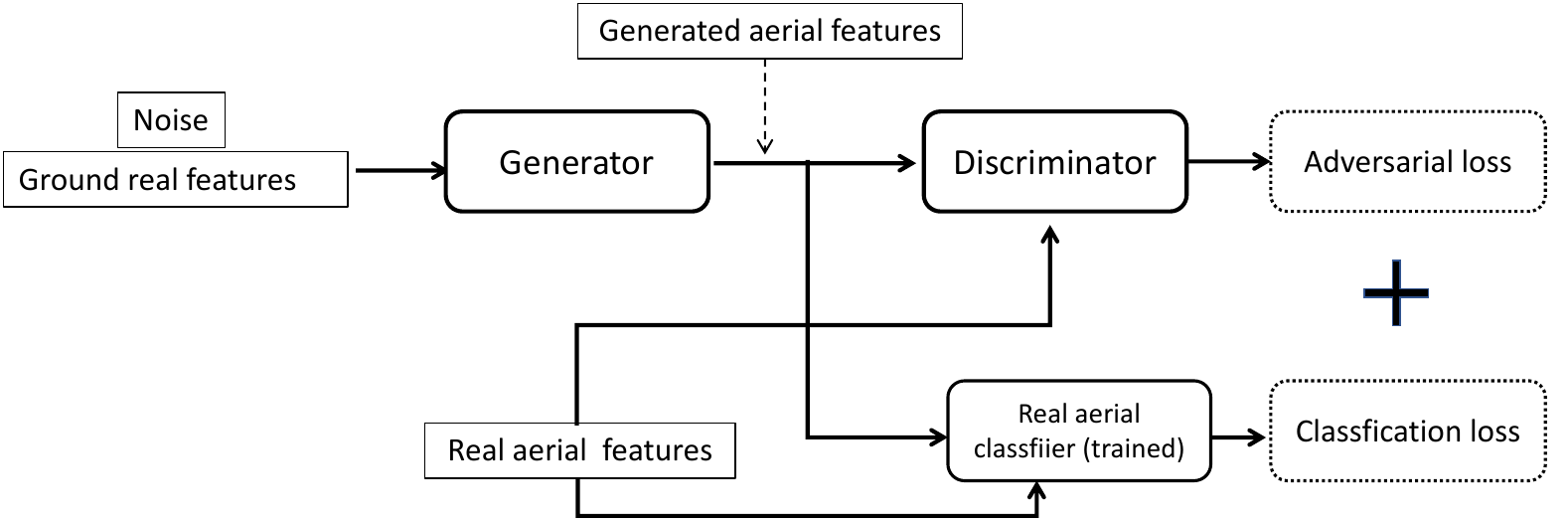}
\end{center}
 \caption{GAN-generated aerial features generation pipeline. Given ground real features, noise and a few real aerial videos, employing adversarial and classification loss, GAN-generated aerial features are generated. }
\label{fig:GAN_pipline}
\end{figure}
Our ultimate goal is to train discriminative action classifiers using GAN-generated features. Although the above objective function generates realistically looking features, it does not guarantee to generate the discriminative features suitable for classification.  To accomplish this, we first train soft-max classifiers using a few available real aerial examples. Finally, to enforce WCGAN-GP to produce discriminative features, we use classification loss computed over the GAN-generated aerial examples given as:
\begin{equation}
  \mathcal{L}_{cl}=-E[log \textit{P} (y_{r_g}|G(z,f_{r_g}); \theta)],
\end{equation}
where \textit{P}($y_{r_g}$$|$G(z,$f_{r_g}$) denotes the probability of correct label prediction of generated examples. Since labels for real ground and GAN-generated aerial examples are the same, we use the labels of real ground videos ($y_{r_g}$) as ground truth. Finally, the overall objective function for GAN-generated aerial examples generation is given by combination of Eq. 2 and Eq. 3. Figure \ref{fig:GAN_pipline} summarize the complete GAN-generated aerial feature generation scheme.


\subsection{Aerial videos classification using Disjoint Multi-Task learning}\label{Sec:DML}
Multitask learning improves the generalization capabilities of the model by effectively learning multiple related tasks.  It has been used in several computer vision problems to learn the joint model such as; simultaneous object detection and segmentation \cite{SimulDetectSeg}, surface normal estimation, and pixel labeling \cite{CrossStitch} and joint pose estimation and action recognition \cite{poseactionrcnn}.
One of the limitations of multitask learning is the requirement of the availability of multiple labels for each task for the \textit{same} data. However, most of existing action datasets do not have such labels and hence restricting multitask learning on these datasets. To address this, recently disjoint multitask learning \cite{DisJMTL} is introduced.  Since, in our approach, the two data sources (games and real)  are different, and secondly, we do not assume any common action classes, this fits well in the context of disjoint multitask learning.

\begin{figure}
\begin{center}
   \includegraphics[width=8.8cm,height=5cm]{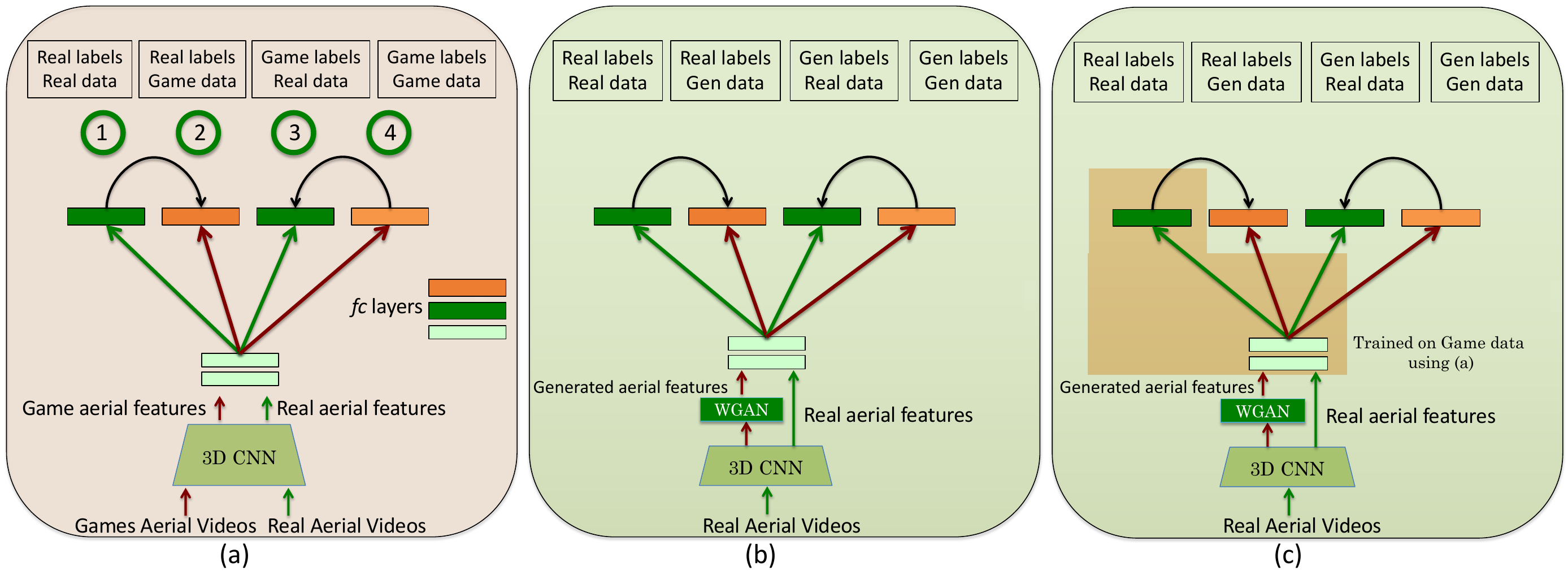}
\end{center}
 \caption{Disjoint multitasking framework for aerial, games and GAN-generated examples.
 (a) Disjoint multitask learning (DML) using games and a few real aerial videos. (b) DML using GAN-generated aerial and real aerial video features (c) DML using real, GAN-generated, and game data. GAN-generated aerial features are abbreviated as Gen data.}
\label{fig:Disjoint Multitasking}
\end{figure}

Figure \ref{fig:Disjoint Multitasking} demonstrates the model overview.
We first compute deep features of a few available real aerial videos and game videos using a 3D convolutional neural network \cite{multifiber, I3D, ResNet3D}. Secondly, we obtain GAN-generated aerial features using the method described in Section \ref{WGAN_Section}. We use two fully connected layers shared between all tasks and one dedicated fully connected layer for each task. Figure \ref{fig:Disjoint Multitasking}.a  shows training using real and game visual features, Figure \ref{fig:Disjoint Multitasking}.b  shows training using real and GAN-generated visual features, and  Figure \ref{fig:Disjoint Multitasking}.c demonstrates training using the real, game and GAN-generated data.  

\noindent\textbf{Joint learning using real and game videos:} We denote the  real and game aerial visual features as 
$r_a$ $\in$ $\mathcal{R}_a$,  $g_a$ $\in$ $\mathcal{G}_a$ respectively.  
Note that we do not assume the type or the number of actions in both datasets to be the same.
We train all four branches of the network for the classification using softmax as a final activation function along with cross-entropy loss.

As shown in Figure \ref{fig:Disjoint Multitasking} (a), we have real and game labels available for real and game data (branch $\textcircled{1}$ and $\textcircled{4}$). However, we do not have real labels for the game data and the game labels for the real data (branch $\textcircled{2}$ and $\textcircled{3}$) due to the disjoint nature of two datasets.
We train different branches of the multi-task framework using the aerial real and game iteratively.
First, we train real and game classification branches ($\textcircled{1}$ and $\textcircled{4}$) for which we have corresponding ground truth labels available. Next, we train branch $\textcircled{2}$ and $\textcircled{3}$ which predicts the real labels for the game data and game labels for real data respectively. However, due to unavailability of labels for $\textcircled{2}$ and $\textcircled{3}$, we use the prediction from $\textcircled{1}$ and $\textcircled{4}$ as a ground truth labels for $\textcircled{2}$ and $\textcircled{3}$ respectively. The overall objective function of the framework is given by:

 \begin{equation}
\begin{aligned}
& \displaystyle\min_{\substack{\bf \Theta}} \sum_{{r_a} \in  \mathcal{R}_a} \overset{{\textcircled{1}}}{\overbrace{\mathcal{L}(y_{r_a}, P (y_{r_a} |r_a ))}}  +  \overset{{\textcircled{3}}}{\overbrace{\mathcal{L}(\hat{y_{g_a}}, P (y_{g_a} |r_a )}}\\ + &  \displaystyle\min_{\substack{\bf \Theta}} \sum_{g_a \in \mathcal{G}_a}   \overset{{\textcircled{2}}}{\overbrace{\mathcal{L}(\hat{y_{r_a}}, P (y_{r_a} |g_a )}}+\overset{{\textcircled{4}}}{\overbrace{\mathcal{L}(y_{g_a}, P (y_{g_a} |g_a ))}}
  \end{aligned}
  \end{equation}
where $y_{r_a}$ and  $P (y_{r_a} |r_a )$ represents ground truth labels and predicted labels of real aerial videos,  $r_a$, $\hat{y_{g_a}}$ are the labels obtained from $\textcircled{4}$ (the layer trained with game ground truth labels) and $P (y_{g_a} |r_a )$ are predicted game action labels for real videos. Similarly, $y_{g_a}$ and $P (y_{g_a} |g_a )$ are  ground truth and predicted action labels of game aerial videos, $\hat{y_{r_a}}$ are obtained from $\textcircled{1}$ (the layer trained with real aerial ground truth labels) and $P (y_{r_a} |g_a )$ is predicted real action labels for game videos. Finally $\bf \Theta$ represents network parameters. We repeat the above procedure for several epochs and fine-tune the parameters on the validation data.

\noindent\textbf{Joint learning using GAN-generated and real videos:}
For joint learning using real and GAN-generated video features, we repeat the same approach as discussed above for real and game videos. Specifically, to obtain an improved action classifier, we feed the network with real and GAN-generated aerial features in an alternating fashion to a disjoint multitask learning framework. Figure \ref{fig:Disjoint Multitasking}(b) illustrates the joint learning using real and GAN-generated video features. Note that since we use real ground features to generate GAN-generated aerial examples (see Figure \ref{fig:GAN_pipline}), we use the labels of real ground videos ($y_{r_g}$) as GAN-generated aerial examples labels, abbreviated as `Gen labels' in the third and fourth branch in Figure \ref{fig:Disjoint Multitasking}(b). As shown in  Figure \ref{fig:Disjoint Multitasking}(a)  and Figure \ref{fig:Disjoint Multitasking}(b), we use the same network architecture for joint learning using games or GAN-generated data.

\noindent\textbf{Joint learning using real, GAN-generated and game videos:}
Finally, we combine all three data types i.e., real, GAN-generated, and game videos in a single framework (see Figure \ref{fig:Disjoint Multitasking}(c)).
In this case, instead of training the network from scratch, we initialize the network with the weights obtained through training using game data. Specifically, networks weights of Figure \ref{fig:Disjoint Multitasking}(a) are used to initialize weights of two backbone ($f_c$) layers and ($f_c$) layer being trained using real data and real labels (the layers are shown in the brown block in Figure \ref{fig:Disjoint Multitasking}(c)). In our case, since the numbers of classes in real and GAN-generated features are the same but that of games are different, we initialize the rest of ($f_c$) layers from scratch. Finally, the network is fine-tuned iteratively by feeding real and GAN-generated data.

\section{Experiments}

The main goal of our experiments is to quantitatively evaluate the proposed approach and analyze the different components. For evaluation, we use two aerial action datasets: UCF-ARG-Aerial \cite{UCFARG} (publicly available) and YouTube-Aerial (will be publicly released). We perform experiments  with and without games and GAN-generated data (Table \ref{table:YouTubeAerial_Overall} and Table \ref{table:UCFARG_Overall}), with and without disjoint multi-task learning (Fig \ref{Table:WithoutDML}). We also performed  K-shot learning experiments (Fig \ref{fig:Graphfordiffnumofvideos}), and analyzed robustness of proposed approach across three different visual features (Table \ref{table:YouTubeAerial_Overall} and Table \ref{table:UCFARG_Overall}). Finally,  action-wise performance on two datasets are given in Table \ref{table:YouTubeAerial_ResultsAL_classwise} and Table \ref{table:UCFARG_ResultsAL_classwise} and confusion matrices for the different components of our approach  are shown in Fig \ref{fig:ConfusionMtrix}.

\subsection{Datasets}
\begin{figure}
\begin{center}
   \includegraphics[width=8.8cm,height=4.5cm]{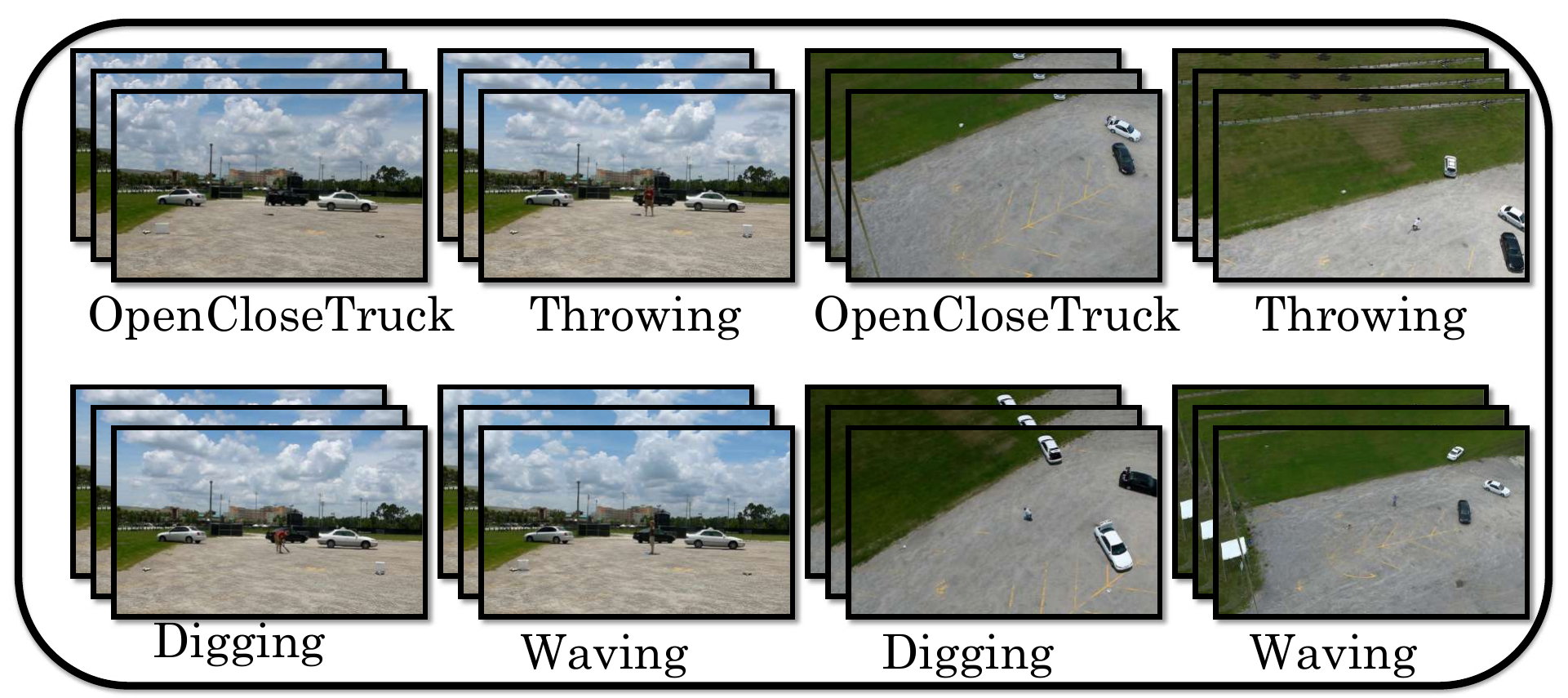}
\end{center}

 \caption{Examples of videos from the UCF-ARG dataset. The first two columns show the videos captured by the ground camera while the last two columns show the same actions captured by a UAV.}
\label{fig:UCFARG}

\end{figure}

\noindent\textbf{UCF-ARG} \cite{UCFARG}:  UCF-ARG dataset contain 10 human actions. This dataset includes: boxing, carrying, clapping, digging, jogging, open-close trunk, running, throwing, walking, and waving. This is a multi-view dataset where videos are collected from an aerial camera mounted on a Helium balloon, ground camera, and rooftop camera. All videos are of high resolution 1920 $\times$ 1080 and recorded at 60fps. The aerial videos contain severe camera shake and large camera motion. On average, each action contains 48 videos. The dataset partition includes 60\% of videos of each action for training, 10\% for validation, and 30\% for testing. Figure \ref{fig:UCFARG} shows some of the videos from the UCF-ARG dataset. Note that the testing experiments are done on the aerial part of the UCF-ARG dataset (named as UCF-ARG-Aerial).

\begin{figure}
\begin{center}
   \includegraphics[width=8.8cm,height=4.5cm]{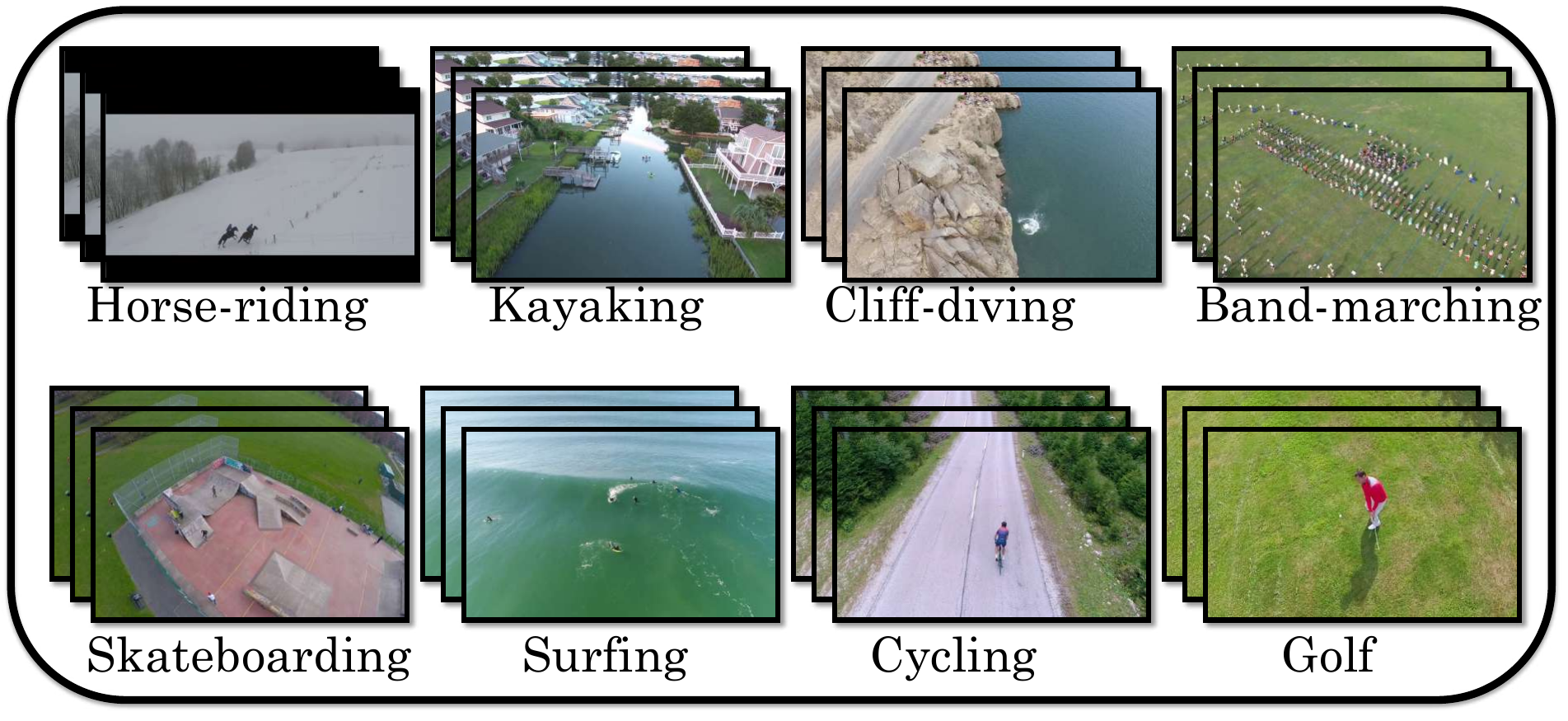}
\end{center}
 \caption{Examples of videos from the YouTube-Aerial dataset. In each video, different human action  is being performed. We aim to automatically recognize human action in these videos when only a few training aerial examples are available.}
\label{fig:Real_Dataset}
\end{figure}

\noindent\textbf{YouTube-Aerial Dataset}: We collect this new dataset ourselves from the drones videos available on YouTube. This dataset contains actions corresponding to eight actions of UCF101 \cite{UCF101}. The actions include band marching, biking,  cliff-diving, golf-swing, horse-riding, kayaking, skateboarding,  and surfing. The videos in this dataset contain large and fast camera motion and aerial videos are captured at variable heights. A few examples of videos in this dataset are shown in Figure \ref{fig:Real_Dataset}. Each action contains 50 videos. Similar to the UCF-ARG dataset, the dataset partition includes 60\%, 10\%, and 30\% of videos for training, validation, and testing respectively.


\subsection{Implementation details}
We use five aerial videos of each action (named as a few available training examples in the above sections).
For visual features computations, we use three recently proposed video features; namely 3D multi-fiber network (MFN-3D) \cite{multifiber}, 3D Inception network (I3D) \cite{I3D}, and 3D residual network (Resnet-3D) \cite{ResNet3D}. Authors in \cite{multifiber} showed that a multi-fiber network provides state-of-the-art results on several competitive datasets and is the order of magnitude faster than several other video features networks. It achieves high computational efficiency by dividing the complex neural network into small lightweight networks. We extract the features (768D) for all videos from the second last layer of the network. I3D features were proposed in \cite{I3D}, where authors suggested a novel technique to inflate 2D ConvNets into 3D and bootstrap 3D filters from 2D filters. We extract the features (1024D) from the global average pooling layer using 128-frame snippets. Experiments are performed using RGB stream only. Similarly, we extract  512 dimension features from the last layer of 3D-Resnet34 \cite{ResNet3D}. 

For disjoint multitask learning, we have two shared fully connected ($f_c$) layers (512 and 256 units respectively). We have four task-specific layers: two $f_c$ layers with the number of units equal to the number of actions in the real dataset and two $f_c$ layers with the number of units equal to the number of actions in the game dataset.  Similarly, for training without DML, we  use only five aerial videos for each action.  We employ three fully connected
($f_c$) layers. The first two  ($f_c$) layers have 512 and 256 units respectively and the third one has the number of units
equal to the number of actions in the dataset. The network is trained using the negative log-likelihood loss. We use the Adam optimizer with learning rate of 0.001, beta1=0.5, and beta2=0.999.

To generate GAN-generated examples, both our generator and discriminator contain four fully connected ($f_c$) layers where the first three $f_c$ layers have Leaky ReLU activation. In the case of the generator, the last  $f_c$ has ReLU activation. The noise vector $z$ (312D) is drawn from unit Gaussian. For all networks, we use Adam optimizer. Since the network is already trained on game data, for joint learning from GAN-generated and game data (Figure \ref{fig:Disjoint Multitasking}(c)), we reduce the weight of loss for the branches being trained on the GAN-generated data. We ran the experiments several times with random initialization of the network ($f_c$ layers) and report the average results.


  \subsection{Experiments Results}

  Table \ref{table:YouTubeAerial_Overall} and Table \ref{table:UCFARG_Overall} demonstrates the experimental results of proposed approach on YouTube-Aerial and UCFARG-Aerial datasets using three visual features \cite{multifiber,I3D,ResNet3D}. All the classification results are on real aerial videos. As compared to the YouTube-Aerial dataset, all visual features have lower classification accuracy on UCF-ARG-Aerial. This is mainly due to two reasons; firstly visual features networks (MFN-3D, I3D, Resnet-3D)  were initially pre-trained on YouTube videos (Kinetics \cite{I3D}, Sports-1M \cite{KarpathyCVPR14} datasets) which are similar to  YouTube-Aerial videos, secondly, UCF-ARG dataset is more challenging due to non-discriminative backgrounds and very small actors size.

  \noindent \textbf{Component-wise accuracy:} In Table \ref{table:YouTubeAerial_Overall} and Table \ref{table:UCFARG_Overall}, trained using `Ground videos' demonstrates classification results when training is done on ground camera videos only. Note that the UCF-ARG dataset contains ground cameras videos for the corresponding aerial action videos.  For the YouTube-Aerial dataset, we use the videos of eight actions from UCF101 ground camera videos. We use ground videos from UCF101 (instead of collecting new ground videos ourselves) because our approach does not require pair-wise correspondence between aerial and ground videos. 
  Note that in our approach ground videos features are only used to generate GAN-generated aerial videos features (see Figure \ref{fig:GAN_pipline}).
  Training using `GAN-generated with DML' demonstrates the experimental results when the network is trained using disjoint multi-task learning employing GAN-generated visual features. Training using `Games with DML' shows the results using game data and finally, training using `Games + GAN-generated with DML' depicts classification results using both game and GAN-generated data.  The experimental results in Table \ref{table:YouTubeAerial_Overall} and Table \ref{table:UCFARG_Overall} show that the proposed approach results in improved aerial action classification. The results emphasize the strength of the proposed approach and suggest that given a few aerial videos (five in our case), games and GAN-generated aerial features can improve the classification accuracy when integrated properly using disjoint multitask learning. 
\begin{table}[htp]
\small\addtolength{\tabcolsep}{-3.0pt}
\footnotesize{}
\centering
\caption{Results of the proposed approach on YouTube-Aerial dataset}\label{table:YouTubeAerial_Overall}
\begin{tabular}{|c|c|c|c|}

\hline
Trained using     & MFN-3D  &  I3D  & Resnet-3D  \\ \hline \hline
Ground videos &49.7 &50.7      & 53.5  
\\ \cline{0-3}
GAN-generated with DML (Fig \ref{fig:Disjoint Multitasking}.b) &64.2 &65.6        & 58.3  \\ \cline{0-3}
Games with DML (Fig \ref{fig:Disjoint Multitasking}.a) &62.9 &64.8        & 56.7  \\ \cline{0-3}
Games + GAN-generated with DML (Fig \ref{fig:Disjoint Multitasking}.c)& \textbf{68.2} & \textbf{67.0}        & \textbf{58.6}  \\ \hline
\end{tabular}

\bigskip
\caption{Results of the proposed approach on UCF-ARG-Aerial dataset for three visual features. The results in Table 1 and Table 2 demonstrate that each component of our approach is important. Both games and GAN-generated visual features are useful for improved aerial classification and combining them further improve the classification results.}\label{table:UCFARG_Overall}
\begin{tabular}{|c|c|c|c|}
\hline
Trained using     & MFN-3D  &  I3D  & Resnet-3D  \\ \hline \hline
Ground videos &21.3  &11.3      & 9.7  
\\ \cline{0-3}
GAN-generated with DML (Fig \ref{fig:Disjoint Multitasking}.b) &32.1 &15.6        & 12.4  \\ \cline{0-3}
Games with DML (Fig \ref{fig:Disjoint Multitasking}.a) &34.4 &\textbf{16.8}        & 13.7  \\ \cline{0-3}
Games + GAN-generated with DML (Fig \ref{fig:Disjoint Multitasking}.c)& \textbf{35.9} &  {16.3}        & \textbf{15.1}  \\ \hline
\end{tabular}

\end{table}

\noindent \textbf{Impact of disjoint multitask learning:} To verify the usefulness of disjoint multitask learning in our approach, 
in Table \ref{Table:WithoutDML}, we show the classification accuracy with and without training using disjoint multitask learning. We use the same experimental settings in both experiments and use the same number (five) of aerial videos. In experiments without DML, we use five aerial videos and in experiments with DML, we use games and GAN-generated data along with five aerial videos. The results demonstrate that integrating games and GAN-generated data through disjoint multitask learning significantly outperforms the training without disjoint multitask learning specifically when the number of available training videos is small.

\begin{figure}[t]
\begin{center}
   \includegraphics[width=8.7cm,height=3.0cm]{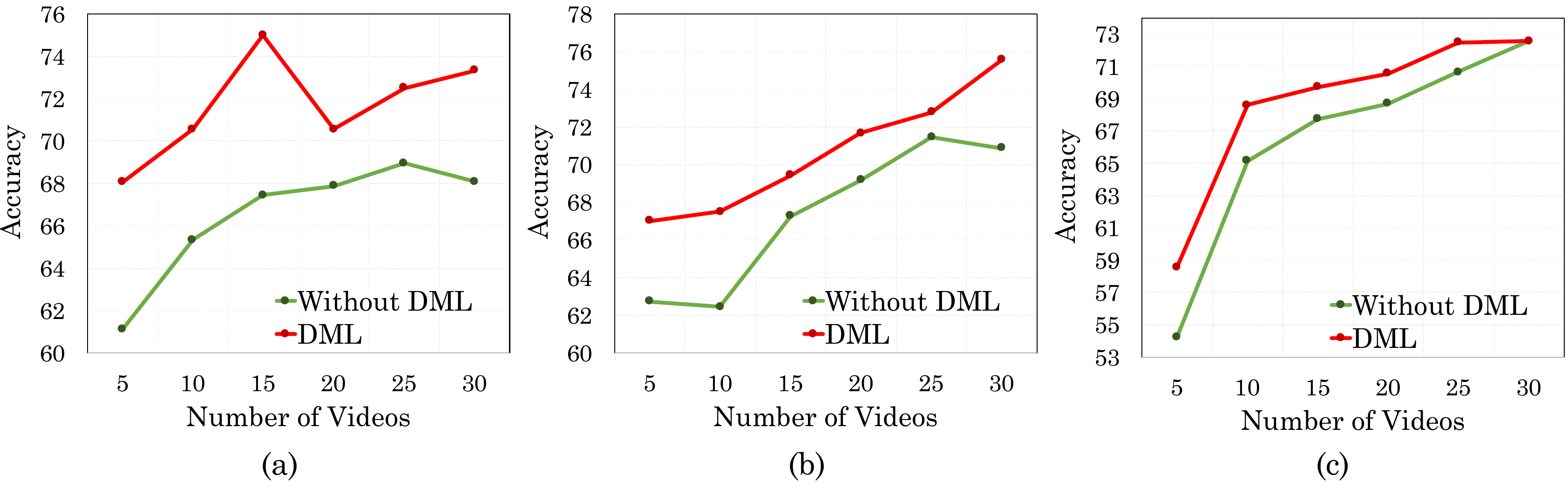}
\end{center} \caption{Accuracy of the proposed approach for the different number of videos. (a), (b), and (c) show the results on YouTube-Aerial dataset with MFN-3D \cite{multifiber}, I3D \cite{I3D} and Resnet-3D \cite{ResNet3D} features respectively. The red curves show the proposed approach while the green curves show the results when training is done without disjoint multitask learning.}
\label{fig:Graphfordiffnumofvideos}
\end{figure}

\tabcolsep=0.5cm
\begin{table}
\caption{Classification results on YouTube-Aerial dataset when training is done without and with employing disjoint multitask learning. In training Without/with disjoint multitask learning,  we have used the same five real aerial videos.} 
\small\addtolength{\tabcolsep}{-7.5pt}
\begin{center}
\footnotesize{
\centering
\begin{tabular}{|c|c|c|c|c|}
\hline
Training & MFN-3D &  I3D  & Resnet-3D \\ \hline \hline
Without Disjoint Multitask Learning & 61.1         & 62.7              &  54.2      \\  
\hline
With Disjoint Multitask Learning & \textbf{68.2}         & \textbf{67.0}             &  \textbf{58.6}      \\ \hline
\end{tabular}}
\end{center}
\label{Table:WithoutDML}
\end{table}
\noindent \textbf{K-shot learning:}
 In  Figure \ref{fig:Graphfordiffnumofvideos}, we demonstrate the accuracy of the proposed approach when training is done using different numbers of aerial videos on the YouTube-Aerial dataset. For the better analysis, we quantitatively compare the proposed approach against training without the disjoint multitask learning framework. The experiments are done for all three visual features. The classification results for the different number of training videos suggest that the proposed approach is not only useful when training data is less but is also beneficial with the increased training data.

\noindent \textbf{Class-wise accuracy:} Table \ref{table:YouTubeAerial_ResultsAL_classwise} and Table \ref{table:UCFARG_ResultsAL_classwise} show the class-wise accuracy of proposed approach using MFN-3D features \cite{multifiber} for UCF-ARG and YouTube-Aerial datasets. For YouTube-Aerial datasets, in five out of eight classes, the proposed approach significantly outperforms the classifier trained only on ground videos. A similar trend can be seen in six out of ten classes of the UCF-ARG dataset. Furthermore, in both datasets,  for the majority of classes, combining GAN-generated and game data either improve the accuracy or keep the best of both.  The proposed approach works better for the actions which have discriminative motion patterns such as Biking, Swing, Kayaking, Carry, Clap, and  Running, etc. However, our approach has limitations for the actions which have strong background scene biases  water or mountains in Diving class) or contain less human body part motion (Skateboarding).
  
\begin{figure}[t]
\begin{center}
   \includegraphics[width=9cm,height=3cm]{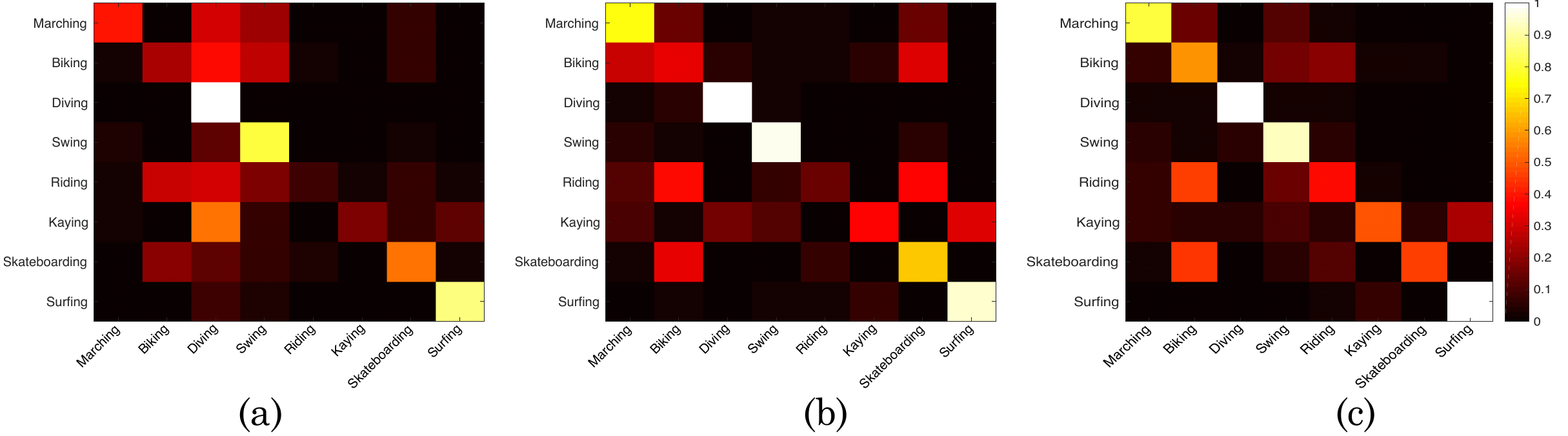}
\end{center}\vspace{-.7cm}
 \caption{This figure shows confusion matrixes averaged over all three visual features (MFN-3D, I3D, and Resnet-3D) on the YouTube-Aerial dataset. Confusion matrixes are (a) for the network trained using ground videos, (b) the network trained using aerial videos without DML,  (c), and the network trained using aerial videos with DML employing both GAN-generated and game data. It can be seen that confusion between actions reduces through DML training.}
\label{fig:ConfusionMtrix}
\end{figure}

\noindent \textbf{Confusion matrices:} Figure \ref{fig:ConfusionMtrix} shows the confusion matrix averaged over all three visual features on the YouTube-Aerial dataset.  The proposed disjoint multi-task learning framework significantly reduced the confusion between different actions as the classification accuracy increases from 51.3 (Figure\ref{fig:ConfusionMtrix}.a) to 59.4 (Figure\ref{fig:ConfusionMtrix}.b) to 64.5 (Figure\ref{fig:ConfusionMtrix}.c).   
 
 \setlength{\tabcolsep}{4pt}
\begin{table*}
    \caption{Quantitative results on YouTube-Aerial dataset. The top row shows class-wise action recognition accuracy on aerial testing videos when trained on done on ground videos.The second, third and forth rows  demonstrate accuracy when training is done using DML employing GAN-generated features, game features and both respectively.}\label{table:YouTubeAerial_ResultsAL_classwise}
    \scriptsize{}\begin{center}
    \label{table:headings}
        \begin{tabular}{@{}llllllllll@{}}
        \hline\noalign{}
        Trained Using                & \rot{\makecell{March-\\ing}}   & \rot{Biking}   & \rot{Diving}   & \rot{Swing}  & \rot{Riding}   & \rot{Kayak}   & \rot{\makecell{Skate-\\board}}   & \rot{Surfing}  & \rot{Avg} \\
        \noalign{}
        \hline
        \noalign{\smallskip}
        Ground Videos & 26.7 & 0 & 100 & 80 & 26.7 & 40 & 53.3 & 73.3 & 49.7 \\
        GAN-generated with DML  & 53.3 & 46.7 & 73.3 & 100 & 40 & 86.7 & 20 & 86.7 & 64.3 \\
        Games with DML & 60 & 73.3 & 86.7 & 100 & 26.7 & 86.7 & 13.3 & 53.3 & 62.9 \\
        Games + GAN-generated with DML& 60 & 73.3 & 86.7 & 100 & 13.3 & 80 & 46.70 & 86.7 & 68.2 \\
        \hline
        \end{tabular}
    \end{center}

\end{table*}
\setlength{\tabcolsep}{1.4pt}

\setlength{\tabcolsep}{4pt}
\setlength{\extrarowheight}{0pt}
\begin{table*}
    \caption{Quantitative results for UCF-ARG dataset. Similar to the Table \ref{table:YouTubeAerial_ResultsAL_classwise}, the top row shows class-wise action recognition accuracy on aerial testing videos when trained on done on ground videos.The second, third and forth rows demonstrate accuracy when training is done using DML employing GAN-generated features, game features and both respectively.}\label{table:UCFARG_ResultsAL_classwise}
    \scriptsize{}\begin{center}
        \begin{tabular}{@{}llllllllllll@{}}
        \hline\noalign{}
        Trained Using                & \rot{Boxing}   & \rot{Carry}   & \rot{Clap}   & \rot{Dig}  & \rot{Jog}   & \rot{Trunk}   & \rot{Run}   & \rot{Throw}  & \rot{Walk}   & \rot{Wave}  & \rot{Avg}\\
        \noalign{}
        \hline
        \noalign{\smallskip}
        Ground Videos                 & 0    & 0    & 0    & 0    & 0    & 0    & 40   & 33.3 & 80   & 60   & 21.33 \\
        GAN-generated with DML         & 60   & 40   & 20   & 13.3 & 13.3 & 58.3 & 26.7 & 6.70 & 13.3 & 66.7 & 31.83 \\
        Games with DML        & 53.3 & 60   & 40   & 00.0 & 6.70 & 83.3 & 40   & 6.70 & 13.3 & 40.0 & 34.33 \\
        Games + GAN-generated with DML & 73.3 & 60.0 & 26.7 & 6.7  & 13.3 & 58.3 & 26.7 & 13.3 & 13.3 & 66.7 & 35.92\\
        \hline
        \end{tabular}
    \end{center}
\end{table*}
\setlength{\tabcolsep}{1.4pt}

\section{Conclusion }
Recently,  low cost and lightweight hardware make drones a good candidate for monitoring human actions. However, training the deep neural network for action recognition needs lots of training examples that are difficult to collect. In this paper, we explore two alternative data sources to obtain more accurate neural network classifiers. To tackle the different types of actions in the game and real action datasets, we propose to use disjoint multitask learning. Our experimental results and thorough analysis demonstrated that game action and GAN-generated examples, when integrated properly, can help to get improved aerial classification accuracy. The future works will aim at spatio-temporal localization of actors in drone videos, which will need attention based deep features.
In our current work, we only use aerial game videos. However, it would be useful to use ground and aerial game videos jointly to learn the transformations between two ground and aerial views. Finally, one of the limitations of drones is their limited battery life. Future work could include designing the algorithms which work on low power devices.

\bibliographystyle{model2-names}
\bibliography{egbib}

\end{document}